\documentclass[letterpaper, 10 pt, conference]{ieeeconf}  % Comment this line out if you need a4paper

\IEEEoverridecommandlockouts                              % This command is only needed if 
\usepackage{amsmath} % assumes amsmath package installed
\usepackage{amssymb}  % assumes amsmath package installed
\usepackage{pifont}
\usepackage{graphicx}
\usepackage{caption}
\usepackage{makecell}
\usepackage{booktabs}
\usepackage{threeparttable}
\usepackage{dirtree}
\usepackage{hyperref}
\usepackage{url}
\usepackage{algpseudocode}
\usepackage{algorithm}
\usepackage{multirow}

\title{\LARGE \bf
ToolTipNet: A Segmentation-Driven Deep Learning Baseline for Surgical Instrument Tip Detection}
\author{Zijian Wu$^{1, \dagger}$, Shuojue Yang$^{2, \dagger}$, Yueming Jin$^{2}$, and Septimiu E. Salcudean$^{1}$% <-this % stops a space
\thanks{$^{1}$ Zijian Wu and Septimiu E. Salcudean are from the Robotics and Control Laboratory (RCL), Department of Electrical and Computer Engineering, the University of British Columbia, Vancouver, Canada.
        {\tt\small zijianwu@ece.ubc.ca}}%
\thanks{$^{2}$ Shuojue Yang and Yueming Jin are from the Department of Biomedical Engineering, National University of Singapore, Singapore.}
\thanks{$^\dagger$ Equal contribution.}
}

\begin{document}

\maketitle
\thispagestyle{empty}
\pagestyle{empty}

%%%%%%%%%%%%%%%%%%%%%%%%%%%%%%%%%%%%%%%%%%%%%%%%%%%%%%%%%%%%%%%%%%%%%%%%%%%%%%%%

% \begin{figure*}[!h]
% \centering{\includegraphics[width=18cm]{Fig1.png}}
% %   \framebox{\parbox{3in}{We suggest that you use a text box to insert a graphic (which is ideally a 300 dpi TIFF or EPS file, with all fonts embedded) because, in an document, this method is somewhat more stable than directly inserting a picture.
% % }}
%   %\includegraphics[scale=1.0]{figurefile}
%   \caption{Inductance of oscillation winding on amorphous
%    magnetic core versus DC bias magnetic field}
%   \label{figurelabel}
% \end{figure*}

\begin{abstract}
In robot-assisted laparoscopic radical prostatectomy (RALP), the location of the instrument tip is important to register the ultrasound frame with the laparoscopic camera frame. A long-standing limitation is that the instrument tip position obtained from the da Vinci API is inaccurate and requires hand-eye calibration. Thus, directly computing the position of the tool tip in the camera frame using the vision-based method becomes an attractive solution. Besides, surgical instrument tip detection is the key component of other tasks, like surgical skill assessment and surgery automation.
However, this task is challenging due to the small size of the tool tip and the articulation of the surgical instrument. Surgical instrument segmentation becomes relatively easy due to the emergence of the Segmentation Foundation Model, i.e., Segment Anything. Based on this advancement, we explore the deep learning-based surgical instrument tip detection approach that takes the part-level instrument segmentation mask as input. Comparison experiments with a hand-crafted image-processing approach demonstrate the superiority of the proposed method on
simulated and real datasets.

% Finally, we address limitations and propose future work to enhance SurgPose.
\end{abstract}

%%%%%%%%%%%%%%%%%%%%%%%%%%%%%%%%%%%%%%%%%%%%%%%%%%%%%%%%%%%%%%%%%%%%%%%%%%%%%%%%

\section{INTRODUCTION}
Surgical instrument tip detection can benefit many applications in Computer-Assisted Intervention (CAI). For example, in RALP, transrectal ultrasound (TRUS) and laparoscope can be registered to each other with a rigid transformation calculated from multiple pairs of TRUS-laparoscope corresponding points~\cite{mohareri2013automatic,wu2023automatic}. This task is approached by using tool tip to poke the tissue surface (air-tissue boundary) where the poking points can be observed and positioned in the ultrasound (US) image. Meanwhile, tip position in the laparoscope frame is estimated using the kinematics chain of da Vinci robot. However, due to the flexibility and backlash of the da Vinci instrument cable, the kinematics data read from API suffers from high inaccuracies that are hard to eliminate by hand-eye calibration~\cite{kalia2021preclinical}.
% The tip of the surgical instrument can be observed in the ultrasound (US) image when the surgeon uses it to poke the tissue surface (air-tissue boundary).
% With multiple pairs of da Vinci-TRUS tool tip locations, the rigid transformation between the da Vinci robot and the TRUS frame can be calculated~\cite{mohareri2013automatic,wu2023automatic}. After this registration, hand-eye (robot-to-camera) calibration of the da Vinci system is necessary since the forward kinematics data read from the API is prone to error~\cite{kalia2021preclinical}. This inaccuracy is problematic to eliminate due to the flexibility and backlash of the da Vinci instrument cable. Furthermore, hand-eye calibration introduces an additional procedure before surgery, which is often time-consuming and cumbersome. 
Driven by these limitations, estimating the position of the instrument tip in the laparoscopic image and then directly registering the laparoscope frame and the TRUS frame becomes an appealing alternative.

For surgical artificial intelligence (AI), surgical instrument tip detection is a fundamental component for many other tasks.
In pose estimation, a line of rendering-based methods has emerged recently and achieved promising performance~\cite{labbe2021single,lu2023markerless,bilic2024gisr}. In recent studies~\cite{yang2025instrument,liang2025differentiable}, tool tip position serving as supervisory signals significantly alleviates the local minimum and thus guides optimization toward the correct pose. For surgical skill assessment, the motion trajectory (and velocity) of the surgical instrument tip plays a crucial role in estimating how well the surgery is performed~\cite{davids2021automated,gruter2023video}. In imitation learning, teaching a robot to do surgery autonomously is challenging. Recent works~\cite{ren2025motion,haldar2025point} represent the robotic action label as 2D trajectories of the gripper keypoints in images/videos. Due to the difficulty of collecting data from the da Vinci robotic system during surgery, detecting surgical instrument tip automatically and learning policy from tool tip motion offers a scalable alternative to achieve surgery autonomy. 

In this work, we aim at vision-based surgical instrument tip detection, which can be regarded as a sub-task of surgical instrument pose estimation (keypoint detection). Deep learning methods~\cite{du2018articulated,kayhan2021deep,li2022multi,ghanekar2025video} have demonstrated superior performance in this task. However, instrument tip detection remains challenging since none of these methods specifically focus on the left and right gripper tips. In this study, we proposed a segmentation-driven approach, \textit{ToolTipNet}, as a strong baseline for surgical tool tip detection. Thanks to the tremendous progress in the segmentation foundation model, surgical instrument segmentation has become a relatively easy task~\cite{kirillov2023segment,wu2025augmenting,ravi2024sam}. A straightforward observation is that instrument tips localize at two special positions of the instrument silhouette.
Inspired by this insight, we try to explore this problem: \textit{Can deep learning model estimate the instrument tip position based on the mask (silhouette) without RGB pixels?} Assuming that the part-level segmentation of the instrument is known, we feed the mask into the deep learning model's backbone and extract the feature pyramid with multiscale information. To enhance the performance, we use the gripper part mask to generate a mask-guided attention map and multiply it by the fused features. Finally, a heatmap head is used to predict the position of the keypoint. 

\section{METHOD}
% This section describes the ToolTipNet architecture and an image processing method for comparison. We also introduce the simulation and real dataset generation method.
\subsection{ToolTipNet Architecture}
\begin{figure}[h]
\centering{\includegraphics[width=0.48\textwidth]{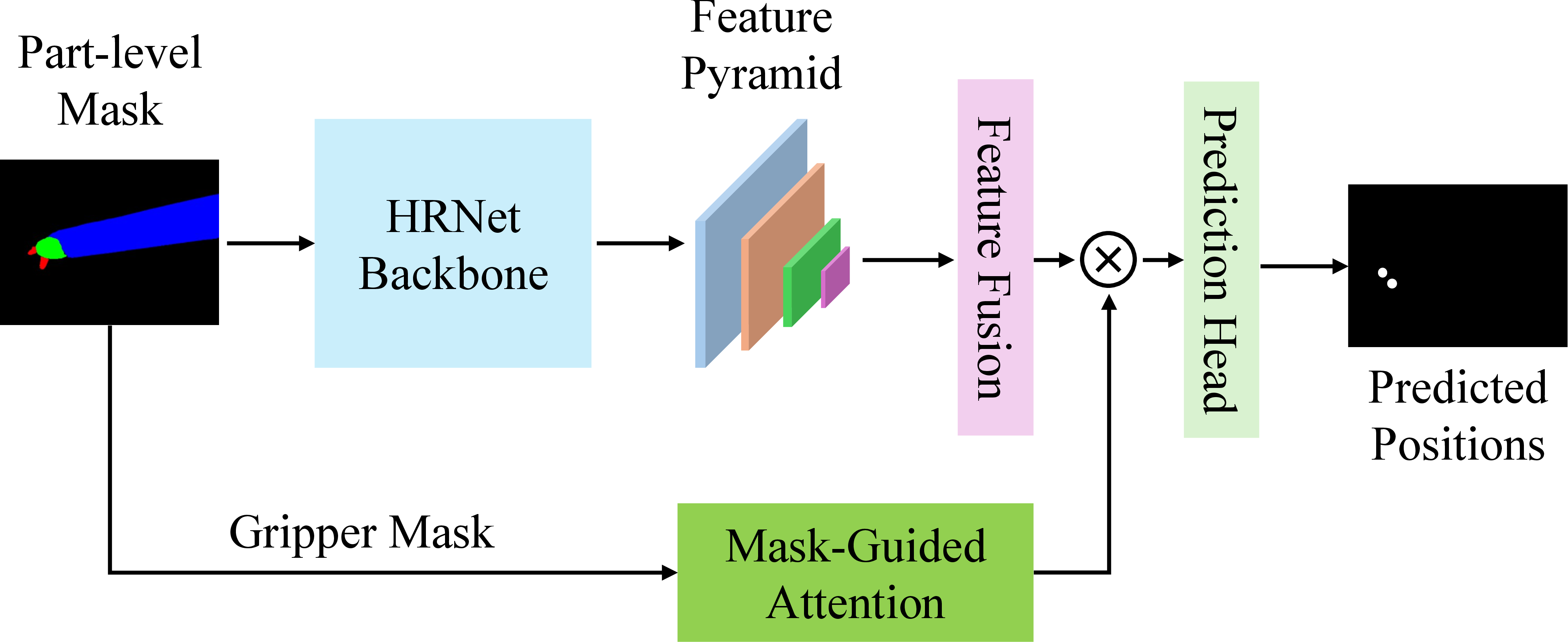}}
  \caption{The architecture diagram of the proposed ToolTipNet.}
  \label{fig5}
\end{figure}

As illustrated in Fig.~\ref{fig5}, the input is the part-level segmentation mask of the instrument and the output is the predicted instrument tips' position. 
% The SAM 2 can generate fairly accurate segmentation mask for the instrument with easy background, so in this study we assume that the accurate surgical instrument mask is known.
In this study, we adopt HRNet~\cite{wang2020deep} as the backbone since it maintains high-resolution representations, which is essential for position-sensitive tasks like keypoint detection. We extract the multi-resolution (1/16, 1/8, 1/4, 1/2) feature pyramid and fuse them into a feature with the shape of (H/4, W/4).
To take advantage of the geometry property of the instrument tip (corner points of the gripper region), we generate a mask-guided attention map and multiply it by the fused feature map. The feature is then fed into the prediction head to output the heatmap of tool tips. 

\subsection{Image-Processing-Based Instrument Tip Detection}
Based on the segmentation mask, hand-crafted features can be used to determine the surgical tool tip. This is the mainstream method in the EndoVis Challenge 2015. In this study, we use an image-processing method~\cite{yang2025instrument} for comparison. We use the Singular Value Decomposition (SVD) to compute the principal axis of the gripper mask. On the principal axis, the pixel farthest from the wrist is considered the instrument tip. 

\subsection{Implementation Details}
In this work, we use pre-trained HRNet-w32 as the backbone based on PyTorch Image Models (timm). For each frame, there is only one instrument with two tips. We adopt loss function $\mathcal L=0.1 \cdot \mathcal L_1+100\cdot \mathcal L_{MSE}$, where $\mathcal L_1$ is the L1 loss of the tool tip coordinates and $\mathcal L_{MSE}$ is the MSE loss of the heatmap. All masks are resized to 512$\times$640 and applied data augmentation (random flip and scale). The model is trained for 100 epochs using Adam optimizer with an initial learning rate of 1e-4 and a batch size of 12. During the training, the learning rate changes according to the Cosine Annealing schedule. All experiments are running on a single NVIDIA RTX 4090 GPU.

\section{EXPERIMENTS AND RESULTS}
\subsection{Datasets \& Metrics}
In our previous work~\cite{yang2025instrument}, we develop a simulator that can place the da Vinci large needle driver CAD model in any pose and generate corresponding part-level semantic mask labels. We generate 9,915 simulated data using this technique.
Using the SurgPose~\cite{wu2025surgpose} data collection method, we also collected some real data clips for evaluation~\cite{yang2025instrument}. We use SAM 2~\cite{ravi2024sam} to extract part-level semantic segmentation mask as the real dataset.
During the training, we use a mixed dataset, which consists of 8,092 simulated masks and 266 real masks. We test the proposed ToolTipNet and SVD-based method on simulation data (1,823 masks) and real data (96 masks). 

As for the metrics, we use 1) Root Mean Squared Error (RMSE) between the prediction and the ground truth instrument tip position and 2) the accuracy of the surgical instrument tip prediction. Here we define that if the RMSE is less than 10 pixels, the prediction is accurate. 
\subsection{Experimental Results}
\begin{figure}[h]
\centering{\includegraphics[width=0.46\textwidth]{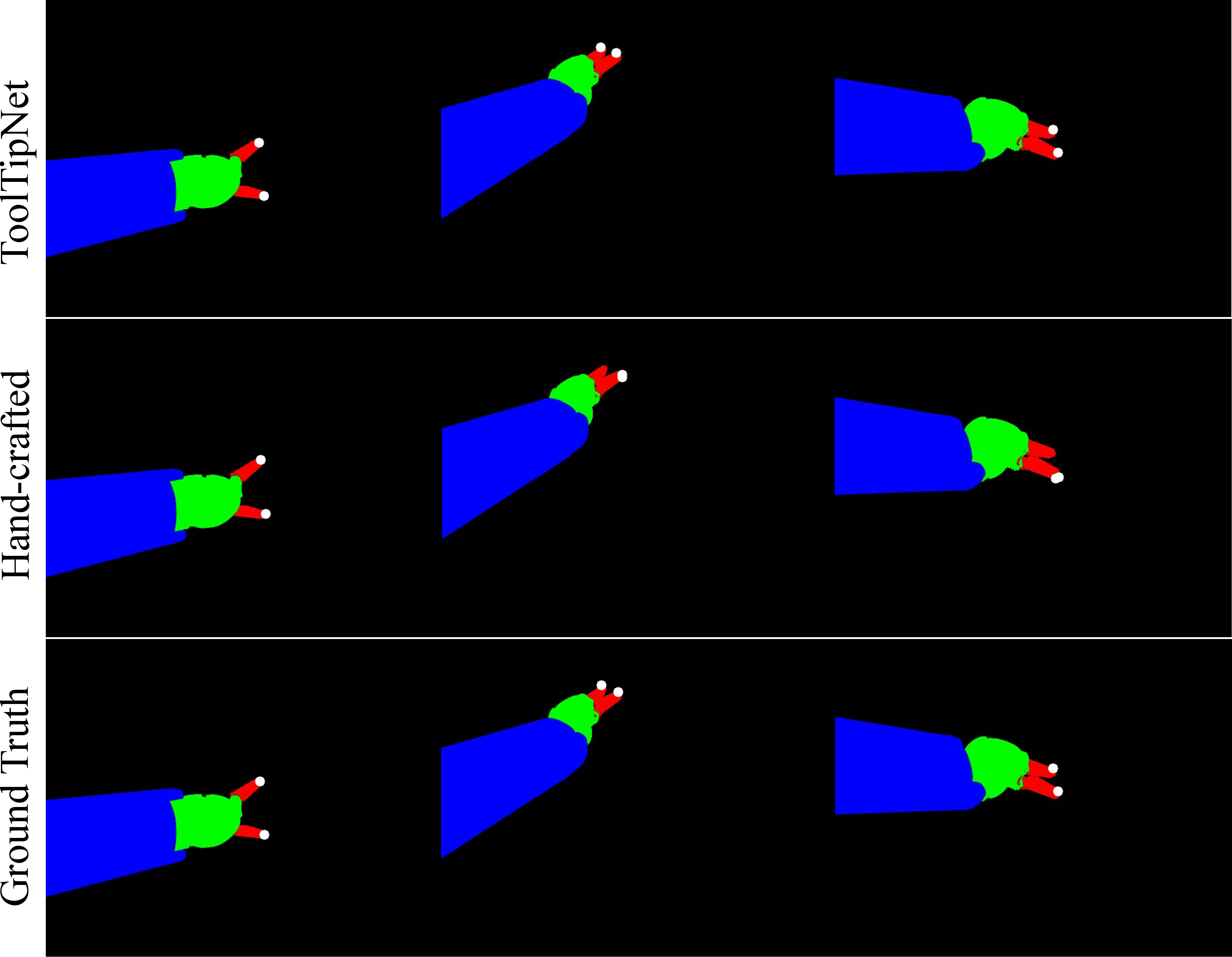}}
  \caption{Visualization of results on simulation mask dataset.}
  \label{fig2}
\end{figure}

\begin{figure}[thb]
\centering{\includegraphics[width=0.46\textwidth]{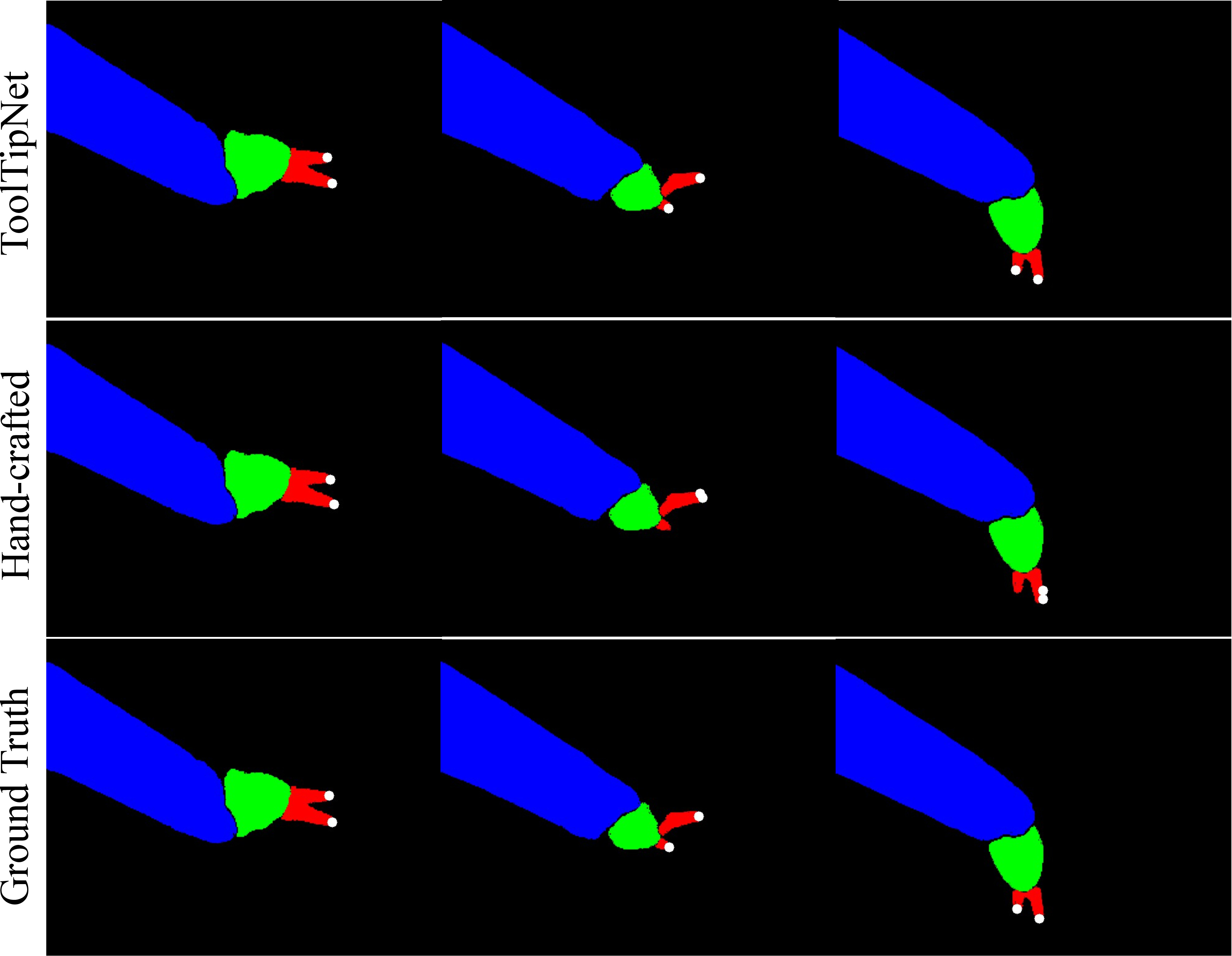}}
  \caption{Visualization of results on real mask dataset.}
  \label{fig3}
\end{figure}

The quantitative results are shown in Table~\ref{table1}. Fig.~\ref{fig2} and~\ref{fig3} are the visualization of the predicted tool tip on the simulated and real datasets. From the empirical observation, the hand-crafted image-processing method often predicts the wrong position when the two grippers are close to each other. A possible reason is that it is difficult to separate the left and right gripper masks into two connected regions. In this situation, SVD cannot compute an accurate principal axis for both grippers and thus cannot predict the correct position. 

\begin{table}[h]
\caption{Quantitative Results on Sim and Real Datasets}
\label{table1}
\begin{center}
\begin{tabular}{c c c c c}
\toprule
 \multirow{2}{*}{Methods} & \multicolumn{2}{c}{Sim. Dataset} & \multicolumn{2}{c}{Real Dataset}\\
 \cmidrule(lr){2-3} \cmidrule(lr){4-5}
 & RMSE $\downarrow$ & Accuracy $\uparrow$ & RMSE $\downarrow$ & Accuracy $\uparrow$ \\
\midrule
% YOLOv8-pose-n & 57.4$\pm$2.04 & 0.8 \\
% YOLOv8-pose-m & 64.1$\pm$1.78  & 3.2 \\
% YOLOv8-pose-l & 65.2$\pm$2.26 & 9.9\\
Yang et al.~\cite{yang2025instrument} & 28.14  & 0.287 & 27.64 & 0.583\\
% \midrule
% ViTPose-S & 48.1$\pm$2.67 & 7.7\\
% ViTPose-B & 56.4$\pm$1.07 & 7.5\\
% ViTPose-L & 43.1$\pm$1.68 & 7.6 \\
\textbf{Ours} & 3.73  & 0.959  & 9.04 & 0.813 \\
% \midrule
% Keypoint RCNN~\cite{he2017mask} & 55.9 $\pm$ & 19.4 \\

\bottomrule
\end{tabular}
\end{center}
\end{table}

% We perform extensive data augmentation, such as mosaic,  to alleviate overfitting.

% Since these baseline methods are originally designed for human pose estimation and are pre-trained on human pose datasets, we adopt the standard metric Object Keypoint Similarity (OKS) defined by COCO. OKS can be calculated using the Euclidean distance between a predicted and a ground truth point, which is passed through an unnormalized Gaussian distribution where the standard deviation corresponds to the square root of the size of the segmentation area multiplied by a per-keypoint constant. 

% The OKS formulation is below:
% $$
% OKS=\frac{\sum_i(exp({\displaystyle\frac{-d_i^2}{2\cdot s^2k_i^2}})\cdot\sigma(v_i>0)}{\sum_i\sigma(v_i>0)}
% $$
% in which $d_i$ are the Euclidean distances between each corresponding ground truth and predicted keypoint. 
% Note that the $k_i$ should not use the value in COCO because the human pose prior is apparently different from the articulated surgical instrument. 
% Similarly to assessing object detection tasks, thresholding the OKS defines matches between the ground truth and estimated keypoints and allows computing precision-recall curves. We adopt the COCO primary metric $mAP$, which is the mean average precision (AP) over multiple OKS thresholds (0.5:0.05:0.95). The results are shown in Table \ref{table_3}. We also show the runtime of each architecture. YOLOv8 and ViTPose can run pose estimation on SurgPose in a real-time manner. The slow inference speed of DeepLabCut is due to the full-size input image.

\section{CONCLUSIONS}
This preliminary study demonstrates the effectiveness of a deep learning baseline ToolTipNet for surgical instrument tip detection given the instrument mask. Future work will focus on reducing the dependency of known segmentation masks. A straightforward method might be to formulate the task as a multi-task learning task, using a foundation model as the backbone along with two branches for segmentation and instrument tip detection, respectively.
% The diverse annotation has the potential for other tasks such as robot joint state estimation.

\section{ACKNOWLEDGMENT}
Zijian Wu is supported by Professor Salcudean's C.A. Laszlo Chair and student support funds from the UBC School of Biomedical Engineering. Computing infrastructure has been provided by the Canada Foundation for Innovation and by the National Science and Engineering Research Council of Canada.

\addtolength{\textheight}{0cm} 


% \section*{ACKNOWLEDGMENT}
%%%%%%%%%%%%%%%%%%%%%%%%%%%%%%%%%%%%%%%%%%%%%%%%%%%%%%%%%%%%%%%%%%%%%%%%%%%%%%%%

% \begin{thebibliography}{99}
\bibliographystyle{IEEEtran}
\bibliography{References}

\end{document}